%% file: root.tex
\documentclass[submission,copyright,creativecommons]{eptcs}
\usepackage{breakurl}             %

\input{utils/packages}
\input{utils/symbols}

\title{A Compositional Sheaf-Theoretic Framework for Event-Based Systems}
\author{Gioele Zardini
\institute{ETH Z\"urich\\
Z\"urich, Switzerland}
\email{gzardini@ethz.ch}
\and
David I. Spivak
\institute{Massachusetts Institute of Technology\\
    Cambridge, MA, USA}
\email{dspivak@gmail.com}
\and 
Andrea Censi
\institute{ETH Z\"urich\\
Z\"urich, Switzerland}
\email{acensi@ethz.ch}
\and
Emilio Frazzoli
\institute{ETH Z\"urich\\
Z\"urich, Switzerland}
\email{efrazzoli@ethz.ch}
}
\def\titlerunning{A Compositional Sheaf-Theoretic Framework for Event-Based Systems}
\def\authorrunning{G. Zardini, D. I. Spivak, A. Censi, and E. Frazzoli}
\begin{document}
\maketitle

\begin{abstract}
A compositional sheaf-theoretic framework for the modeling of complex event-based systems is presented. We show that event-based systems are machines, with inputs and outputs, and that they can be composed with machines of different types, all within a unified, sheaf-theoretic formalism. We take robotic systems as an exemplar of complex systems and rigorously describe actuators, sensors, and algorithms using this framework.
\end{abstract}

\input{chapters/introduction}
\input{chapters/background}
\input{chapters/event}

\clearpage
\input{chapters/examples}

\clearpage
\input{chapters/conclusion}

\nocite{*}
\bibliographystyle{eptcs}
\bibliography{generic}
\end{document}

%% file: utils/packages.tex
\usepackage{mathtools}
\usepackage{subfig}
\usepackage{comment}
\usepackage{amsthm}
\usepackage[varg,bigdelims]{newpxmath}
\usepackage[acronym]{glossaries}
\usepackage{tikz}
\usepackage{stmaryrd}
\usepackage{enumitem}
\usepackage[capitalize]{cleveref}
\usepackage{xcolor}
  \usetikzlibrary{ 
  	cd,
  	math,
  	decorations.markings,
  	decorations.pathmorphing,
		decorations.pathreplacing,
  	arrows.meta,
  	calc,
  	fit,
  	quotes,
  	positioning
  }
  
\tikzcdset{arrow style=tikz, diagrams={>=To}}

\tikzstyle{block} = [draw, rectangle, minimum height=2.5em, minimum width=3.5em]

\tikzset{fcname/.store in =\fcname, fcname=nan}
\tikzset{funame/.store in =\funame, funame=nan}
\tikzset{rcname/.store in =\rcname, rcname=nan}
\tikzset{runame/.store in =\runame, runame=nan}
\tikzset{feedback/.store in =\feedback, feedback=0}
\tikzset{
   DP/.style={%
      label/.style={
         font=\everymath\expandafter{\the\everymath\scriptstyle},
         inner sep=0pt,
         node distance=2pt and -2pt},
      semithick,
      node distance=1 and 1,
      rconn/.style={color=white,opacity=0.0,postaction={decorate}, shorten <=3.2pt, shorten >= 0.8,
      decoration={markings, 
      mark= at position 0 with {
               \coordinate (a);
      },
      mark=at position .5 with
      {
              \ifthenelse{\equal{\feedback}{1}}{\def\angleOut{90}\def\angleIn{90}}{\def\angleOut{0}\def\angleIn{180}}    
              \coordinate (b);
              \draw[dashed,red,opacity=1.0] (a) to[out=\angleOut,in=\angleIn,looseness=2] 
              node[pos=0.25,above=0.5mm,red,opacity=1,fill=white,inner sep=1pt,outer sep=1pt]{\scriptsize{\rcname}} (b);
      },
      mark= at position 1 with 
      {
              \ifthenelse{\equal{\feedback}{1}}{\def\angleOut{0}\def\angleIn{0}}{\def\angleOut{180}\def\angleIn{0}} 
              \ifthenelse{\equal{\feedback}{1}}{\def\symbol{\succeq}}{\def\symbol{\preceq}} 
              \coordinate (c);
              \draw[green,opacity=1.0] (c) to[out=\angleOut,in=\angleIn,looseness=2]
              node[pos=0.25,above=0.5mm,green,opacity=1,fill=white,inner sep=1pt,outer sep=1pt]{\scriptsize{\fcname}} (b){}; %
              \node[draw,circle,inner sep=0.5pt,color=black,fill=white,opacity=1.0] at (b) (nodepreceq) {$\symbol$}; 
      }
      }},
      runconn/.style={color=red,dashed,postaction={decorate},
      decoration={markings,
      mark= at position 1 with {
              \coordinate (a);
              \draw[red,opacity=1.0,dashed] ($(a)+(0.05,0)$) --++ (1.5,0) node[right,pos=0.95]{\scriptsize{\runame}};}
      }
      },
      funconn/.style={color=white,postaction={decorate},
      decoration={markings,
      mark= at position 0 with {
      \coordinate (a);
      \draw[green] ($(a)+(-0.05,0)$) -- ($(a)+(-1,0)$) node[left]{\scriptsize{\funame}};}
      }
      },
      execute at begin picture={\tikzset{
         x=\dpx, y=\dpy,
         every fit/.style={inner xsep=\dpx, inner ysep=\dpy}}}
      },
   dpx/.store in=\dpx,
   dpx = 1.5cm,
   dpy/.store in=\dpy,
   dpy = 1.75ex,
   dp port sep/.store in=\dpportsep,
   dp port sep=2,
   dp port length/.store in=\dpportlen,
   dp port length=4pt,
   dp min width/.store in=\dpminwidth,
   dp min width=1.5cm,
   dp rounded corners/.store in=\dpcorners,
   dp rounded corners=2pt,
   dp small/.style={dp port sep=1, dp port length=2.5pt, dpx=.4cm, dp min width=.4cm, dpy=.7ex},
   dp/.code 2 args={%
      \pgfmathsetlengthmacro{\dpheight}{\dpportsep * (max(#1,#2)+1) * \dpy}
      \pgfkeysalso{draw,minimum height=\dpheight,minimum width=\dpminwidth,outer sep=0pt,
         rounded corners=\dpcorners,thick,
         prefix after command={\pgfextra{\let\fixname\tikzlastnode}},
         append after command={\pgfextra{\draw
            \ifnum #1=0{} \else foreach \i in {1,...,#1} {
               ($(\fixname.north west)!{\i/(#1+1)}!(\fixname.south west)$) +(0,0) node[solid,left,circle,color=green,draw,fill=green,scale=0.3] {} coordinate (\fixname_fun\i) -- +(0,0) coordinate (\fixname_fun\i')}\fi %
            \ifnum #2=0{} \else foreach \i in {1,...,#2} {
               ($(\fixname.north east)!{\i/(#2+1)}!(\fixname.south east)$) +(0,0) coordinate (\fixname_res\i') -- +(0,0) node[solid,right,circle,color=red,draw,fill=red,scale=0.3] {} coordinate (\fixname_res\i)}\fi;
         }}}
   },
   dp name/.style={append after command={\pgfextra{\node[label=center] at (\fixname) {\textbf{#1}};}}}
   }
   
   \tikzset{
   oriented WD/.style={%
      every to/.style={out=0,in=180,draw},
      label/.style={
         font=\everymath\expandafter{\the\everymath\scriptstyle},
         inner sep=2pt,
         node distance=2pt and -2pt},
      semithick,
      node distance=1 and 1,
      decoration={markings, mark=at position .5 with {\arrow{stealth};}},
      ar/.style={postaction={decorate}},
      execute at begin picture={\tikzset{
         x=\bbx, y=\bby,
         every fit/.style={inner xsep=\bbx, inner ysep=\bby}}}
      },
   bbx/.store in=\bbx,
   bbx = 1.5cm,
   bby/.store in=\bby,
   bby = 1.75ex,
   bb port sep/.store in=\bbportsep,
   bb port sep=2,
   bb port length/.store in=\bbportlen,
   bb port length=4pt,
   bb min width/.store in=\bbminwidth,
   bb min width=1cm,
   bb rounded corners/.store in=\bbcorners,
   bb rounded corners=2pt,
   bb small/.style={bb port sep=1, bb port length=2.5pt, bbx=.4cm, bb min width=.4cm, bby=.7ex},
   bb/.code 2 args={%
      \pgfmathsetlengthmacro{\bbheight}{\bbportsep * (max(#1,#2)+1) * \bby}
      \pgfkeysalso{draw,minimum height=\bbheight,minimum width=\bbminwidth,outer sep=0pt,font=\bfseries,inner sep=5pt,
         rounded corners=\bbcorners,thick,
         prefix after command={\pgfextra{\let\fixname\tikzlastnode}},
         append after command={\pgfextra{\draw
            \ifnum #1=0{} \else foreach \i in {1,...,#1} {
               ($(\fixname.north west)!{\i/(#1+1)}!(\fixname.south west)$) -- +(-2*\bbportlen,0) coordinate (\fixname_input\i)}\fi %
            \ifnum #2=0{} \else foreach \i in {1,...,#2} {
               ($(\fixname.north east)!{\i/(#2+1)}!(\fixname.south east)$) -- +(2*\bbportlen,0) coordinate (\fixname_out\i)}\fi;
         }}}
   },
   bb name/.style={append after command={\pgfextra{\node[anchor=north] at (\fixname.north) {#1};}}}
}
   
\theoremstyle{definition}
\newtheorem{theorem}{Theorem}[section]
  \newtheorem{proposition}[theorem]{Proposition}

  \newtheorem{definition}[theorem]{Definition}

  \newtheorem*{axiom*}{Axiom}
  \newtheorem{example}[theorem]{Example}

  \theoremstyle{remark}
  \newtheorem{remark}[theorem]{Remark}

\DeclareSymbolFont{stmry}{U}{stmry}{m}{n}
\DeclareMathSymbol\fatsemi\mathop{stmry}{"23}

\DeclareFontFamily{U}{mathx}{\hyphenchar\font45}
\DeclareFontShape{U}{mathx}{m}{n}{
      <5> <6> <7> <8> <9> <10>
      <10.95> <12> <14.4> <17.28> <20.74> <24.88>
      mathx10
      }{}
\DeclareSymbolFont{mathx}{U}{mathx}{m}{n}
\DeclareFontSubstitution{U}{mathx}{m}{n}
\DeclareMathAccent{\widecheck}{0}{mathx}{"71}

%% file: utils/symbols.tex
\newcommand{\Cat}[1]{\mathsf{#1}}%
\newcommand{\cat}[1]{\mathcal{#1}}%
\newcommand{\clock}[1]{\mathsf{Clock}_{#1}}

\newcommand{\cont}[1]{\text{Cnt}_{#1}}

\newacronym{abk:cps}{CPS}{Cyber-physical system}

\newcommand{\dir}{\mathrm{dir}}

\newcommand{\event}[1]{\text{Ev}_{#1}}

\newcommand{\eventcat}{\Cat{Event}}

\newcommand{\fcnt}{f^\text{cnt}}
\newcommand{\fin}{f^\text{in}}
\newcommand{\fevt}{f^\text{evt}}
\newcommand{\fdyn}{f^\text{dyn}}
\newcommand{\fout}{f^\text{out}}

\newcommand{\identity}[1]{\text{id}_{#1}}
\newcommand{\Int}{\Cat{Int}}

\newcommand{\lcont}[1]{\text{LCnt}_{#1}}

\newcommand{\mach}{\Cat{Mach}}
\newcommand{\op}[1]{{#1}^\text{op}}
\newcommand{\tn}[1]{\textnormal{#1}}
\newcommand{\phase}[1]{\text{Ph}_{#1}}
\newacronym{abk:poset}{poset}{partially ordered set}

\newcommand{\Pair}[2]{\left( {#1},{#2}\right)}
\newcommand{\triple}[3]{\left( #1,#2,#3\right)}

\newcommand{\proofref}[1]{\ifthenelse{\boolean{submission}}{}{The proof can be found in \cref{#1}.}}

\newcommand{\rest}[3]{{#1}|_{[#2,#3]}}
\newcommand{\Rgeq}{\mathbb{R}_{\geq 0}}
\newcommand{\rotatedpullbacksign}{\rotatebox[origin=c]{-45}{$\lrcorner$}}
\newcommand{\rotatedpullback}{\arrow[dd, phantom, "\rotatedpullbacksign", very near start]}

\newcommand{\set}{\Cat{Set}}
\newcommand{\sheaf}[1]{\widetilde{#1}}

\DeclareMathOperator{\ob}{Ob}

\newcommand{\rr}{\mathbb{R}}
\renewcommand{\ss}{\subseteq}

\newcommand{\inv}{^{-1}}

\newcommand{\qqand}{\qquad\text{and}\qquad}
\newcommand{\upd}{^{\tn{upd}}}
\newcommand{\rdt}{^{\tn{rdt}}}

\newcommand{\then}{\fatsemi}

\newcommand{\dnote}[1]{\textnormal{\color{blue}David says: #1}}

\newcommand{\anote}[1]{\textnormal{\color{green!50!black}Andrea says: #1}}

\newcommand{\To}[1]{\xrightarrow{#1}}
\newcommand{\From}[1]{\xleftarrow{#1}}

%% file: chapters/introduction.tex
\section{Introduction}
\label{sec:intro}
This paper presents a unified modeling framework for event-based systems, focusing on the standard example of cybernetic systems. Specifically, we present event-based systems as machines, showing how to compose them in various ways and how to describe diverse interactions (between systems that are continuous, event-based, synchronous, etc). We showcase the efficacy of our framework by using it to describe a robotic system, and we demonstrate how apparently different robotic components, such as actuators, sensors, and algorithms, can be described within a common sheaf-theoretic formalism. 

\paragraph{Related Work.} \glspl{abk:cps}, first mentioned by Wiener in 1948~\cite{Wiener1948}, consist of an \emph{orchestration of computers and physical systems}~\cite{Lee2015} allowing the solution of problems which neither part could solve alone. \glspl{abk:cps} are complex systems including physical components, such as actuators, sensors, and computer units, and software components, such as perception, planning, and control modules. Starting from the first mention of \glspl{abk:cps}, their formal description has been object of studies in several disciplines, such as control theory~\cite{Branicky1998,tabuada2004,Dimarogonas2011,Heemels2012}, computer science~\cite{Henzinger2000,Platzer2008,Platzer2015,Lee2016,Lee2018}, and applied category theory~\cite{Ames2006,Schultz2019,Schultz2020}. 

Traditionally, control theorists approached \glspl{abk:cps} through the study of hybrid systems, analyzing their properties and proposing strategies to optimally control them.  In~\cite{Branicky1998}, researchers identify phenomena arising in real-world hybrid systems and introduce a mathematical model for their description and optimal control. In~\cite{tabuada2004}, a compositional framework for the abstraction of discrete, continuous, and hybrid systems is presented, proposing constructions to obtain hybrid control systems. Furthermore,~\cite{Dimarogonas2011} defines event-driven control strategies for multi-agent systems, and ~\cite{Heemels2012} proposes an introduction to event- and self-triggered control systems, which are proactive and perform sensing and actuation when needed.

On the other hand, computer scientists focused on the study of verification techniques for \glspl{abk:cps}. In~\cite{Henzinger2000}, the researcher introduces a comprehensive theory of hybrid automata, and focuses on tools for the reliability analysis of safety-critical \glspl{abk:cps}. Similarly,~\cite{Platzer2008,Platzer2015} develop a differential dynamic logic for hybrid systems and introduce deductive verification techniques for their safe operation. Furthermore,~\cite{Lee2016,Lee2018} underline that to achieve simplicity and understandability, \emph{clear, deterministic modeling semantics} have proven valuable in modeling \glspl{abk:cps}, and suggest the use of modal models, ensuring that models are used within their validity regime only (e.g. discrete vs. continuous).

In 2006, ~\cite{Ames2006} proposes the first category-theoretic framework for the study of hybrid systems, defining the category of hybrid objects and applying it to the study of bipedal robotic walking. Finally, ~\cite{Speranzon2018,Schultz2019,Schultz2020} present temporal type theory and machines, which are shown to be able to describe discrete and continuous dynamical systems, and which lay the foundation for this work.

\paragraph{Organization of the Paper.}\cref{sec:back} recalls the theory of sheaves and machines presented in~\cite{Schultz2020} and includes explanatory examples. \cref{sec:eventbased} shows how to model event-based systems within this framework, and provides practical tools and examples. \cref{sec:examples} showcases the properties of the defined framework, by employing it to model the feedback control of a flying robot.

\paragraph{Acknowledgements}
We would like to thank Dr. Paolo Perrone for the fruitful discussions. DS acknowledges support from AFOSR grants FA9550-19-1-0113 and FA9550-17-1-0058.

%% file: chapters/background.tex
\section{Background}
\label{sec:back}
The reader is assumed to be familiar with category theory. An extended, self-contained version of this article is reported in~\cite{Zardini2020}. In this section, we review the material needed to present Section~\ref{sec:eventbased}. This paper builds on the theory presented in~\cite{Schultz2019,Schultz2020}.
Let $\Rgeq$ denote the linearly ordered poset of non-negative real numbers. For any $a\in \Rgeq$, let
\begin{equation*}
    \begin{split}
        \phase{a}\colon \Rgeq &\to \Rgeq \\
        \ell&\mapsto a+\ell.
    \end{split}
\end{equation*} 
denote the translation-by-$a$ function.

\begin{definition}[Category of continuous intervals $\Int$]
\label{def:categoryint}
The \emph{category of continuous intervals $\Int$} is composed of:
\begin{itemize}
    \item Objects: $\ob(\Int)\coloneqq \Rgeq$. We denote such an object by $\ell$ and refer to it as a \emph{duration}.
    \item Morphisms: Given two durations $\ell$ and $\ell'$, the set $\Int(\ell,\ell')$ of morphisms between them is 
    \begin{equation*}
        \Int(\ell,\ell')\coloneqq\{ \phase{a} \mid a \in \Rgeq \text{ and }a+\ell\leq \ell'\}.
    \end{equation*}
    \item The identity morphism on $\ell$ is the unique element  $\identity{\ell}\coloneqq\phase{0}\in\Int(\ell,\ell).$
    \item Given two morphisms $\phase{a}\colon \ell\to \ell'$ and $\phase{b}\colon\ell'\to \ell''$, we define their composition as $\phase{a}\then \phase{b}= \phase{a+b}\in\Int(\ell,\ell'')$.
\end{itemize}
\end{definition}

We often denote the interval $[0,\ell]\ss\rr$ by $\tilde\ell$. A morphism $\phase{a}$ can be thought of as a way to include $\tilde\ell'$ into $\tilde\ell$, starting at $a$, i.e.\ the subinterval $[a,a+\ell']\ss[0,\ell]$.%
\footnote{$\Int$ is not just the category of intervals $[a,b]$ with inclusions, which we temporarily call $\Int'$; in particular $\Int'$ is a poset, whereas $\Int$ is not. For experts: $\Int'$ is the twisted arrow category of the poset $(\rr,\leq)$, whereas $\Int$ is the twisted arrow category of the monoid $(\rr,+,0)$ as a category with one object.}
\begin{proposition}
\label{prop:intcat}
$\Int$ is indeed a category: It satisfies associativity and unitality. \proofref{app:intcat}
\end{proposition}

\begin{definition}[$\Int$-presheaf]
\label{def:intsheaf}
An \emph{$\Int$-presheaf} $A$ is a functor
\begin{equation*}
    A\colon \op{\Int} \to \set,
\end{equation*}
where $\set$ is the category of sets and functions. Given any duration $\ell$, we refer to elements $x\in A(\ell)$ as \emph{length-$\ell$ sections (behaviors)} of $A$. For any section $x\in A(\ell)$ and any map $\phase{a}\colon \ell'\to \ell$ we write $\rest{x}{a}{a+\ell'}$ to denote its \emph{restriction} $A(\phase{a})(x)\in A(\ell')$. 
\end{definition}

\begin{definition}[Compatible sections]
If $A$ is an $\Int$-presheaf, we say that sections $a\in A(\ell)$ and $a'\in A(\ell')$ are \emph{compatible} if the right endpoint of $a$ matches the left endpoint of $a'$, i.e.:
\begin{equation*}
    \rest{a}{\ell}{\ell}=\rest{a'}{0}{0}.
\end{equation*}
\end{definition}

\begin{definition}[$\Int$-sheaf]
An $\Int$-presheaf 
\begin{equation*}
    P\colon \op{\Int}\to \set.
\end{equation*}
is called an $\Int$-\emph{sheaf} if, for all $\ell,\ell'$ and compatible sections $p\in P(\ell)$, $p'\in P(\ell')$ (i.e., with $\rest{p}{\ell}{\ell}=\rest{p'}{0}{0}$), there exists a unique $\bar{p}\in P(\ell+\ell')$ such that 
\begin{equation*}
    \rest{\bar{p}}{0}{\ell}=p \qqand \rest{\bar{p}}{\ell}{\ell+\ell'}=p'.
\end{equation*}
Morphisms of $\Int$-sheaves are just morphisms of their underlying $\Int$-presheaves. We denote the category of $\Int$-sheaves as $\text{Shv}(\Int)\coloneqq\sheaf{\Int}$.
\end{definition}

\begin{example}[Initial and terminal objects in $\sheaf{\Int}$]\label{ex_init_term_behavior_types}
The terminal object in $\sheaf{\Int}$ is called $1$, and it assigns to each interval $\ell$ the one-element set $\{1\}$. Similarly $\varnothing\in\sheaf{\Int}$ is the initial object and sends each interval $\ell$ to the empty set $\varnothing$.
\end{example}

\begin{example}[Period-$d$ clock]
For any $d>0$ define $\clock{d}$ to be the presheaf with
\begin{equation*}
    \clock{d}(\ell)\coloneqq\big\{\{t_1,\ldots,t_n\}\ss\tilde{\ell}\mid t_1<d, \ell-t_n<d, \tn{ and } t_{i+1}-t_i=d \tn{ for all }1\leq i\leq n{-}1\big\}.
\end{equation*}
Note that $(\ell/d)-1\leq n\leq \ell/d$. We denote an element of $\mathrm{Clock}_d(\ell)$ by $\phi=\{t_1,\ldots,t_n\}\ss\tilde{\ell}$; it's the set of ``ticks'' of the clock, spaced $d$-apart. Given $\phi$ and $\phase{a}\colon\ell'\to \ell$, the restriction is given by taking those ``ticks'' that are in the smaller interval: \[\clock{d}(\phase{a})(\phi)=\rest{\phi}{a}{a+\ell'}\coloneqq \phi\cap\tilde\ell'.\]
\end{example}

\begin{definition}[Machine]\label{def.mach}
Let $A,B \in \sheaf{\Int}$. An \emph{$(A,B)$ machine} is a span in $\sheaf{\Int}$
\begin{center}
\input{chapters/tikz/20_machine}
\end{center}
Equivalently, it is a sheaf $C$ together with a sheaf map $f \colon C\to A\times B$.%
\footnote{Technically, we identify spans $(C,f)$ and $(C',f')$ if there is an isomorphism $i\colon C\to C'$ with $i\then f'=f$.}
We refer to $A$ as the \emph{input sheaf}, to $B$ as the \emph{output sheaf}, and to $C$ as the \emph{state sheaf}.
\end{definition}

\begin{remark}\label{rem.det_tot_inert}
\cref{def.mach} is symmetric, whereas machines are generally considered causal: inputs affect later outputs. This is captured formally by the notion of being \emph{total, deterministic, and inertial}, defined in \cite{Schultz2020}. Totalness (resp.\ determinism) means that given $c\in C(\ell)$ and $a'\in A(\ell')$ with $\rest{\fin(c)}{0}{\ell}=a$, there is at least one (resp.\ at most one), $c'\in C(\ell)$ such that $\rest{c'}{0}{\ell}=c$ and $\fin(c')=a'$. In other words, there is a unique way that the internal behavior can accommodate any input coming in. A machine is $\varepsilon$-inertial if its internal state on an interval $[a,b]$ determines its output on $[a,b+\varepsilon]$.

In this paper, all of our machines will be total and deterministic; however we will not mention this explicitly. Each can also be made $\varepsilon$-inertial by composing it (\cref{def:composition}) with an $\varepsilon$-delay (\cref{ex:delay}). A result of \cite{Schultz2020} says that if all the machines in a network are deterministic, total, and inertial, then their composite is as well (\cref{def:composition}). 
\end{remark}

\begin{example}[Continuous dynamical system]
\label{ex:cds}
For Euclidean spaces $A,B$, an \emph{$(A,B)$-continuous dynamical system (CDS)} consists of
\begin{itemize}
    \item $S=\rr^n$, called the \emph{state space}.
    \item The \emph{dynamics} $\dot{s}=\fdyn(s,a)$, for any $a\in A$, $s\in S$, and smooth function $\fdyn$.
    \item The \emph{readout} $b=f\rdt(s)$, with $b\in B$ and smooth function $f\rdt$.
\end{itemize}
We can write this as a machine
\begin{center}
\input{chapters/tikz/20_cds}
\end{center}
where the apex is given by 
\begin{equation*}
D(\ell)=\{(a,s,b)\in C^1(A) \times S \times C^1(B) \mid \dot{s}=\fdyn(a,s) \tn{ and } b=f\rdt(s)\},
\end{equation*}
and $\fin$ and $\fout$ are the projections $(a,s,b)\mapsto a$ and $(a,s,b)\mapsto b$. Note that $C^1(A),C^1(B)$ represent the continuously differentiable functions on $A$ and $B$, respectively. 
\end{example}

\begin{example}[$\varepsilon$-delay]
\label{ex:delay}
Given any sheaf $A$ and positive real $\varepsilon>0$, we define the \emph{$A$-type $\varepsilon$-delay machine} $\mathrm{Del}^A_\varepsilon$ to be the span $A\From{\fin} A_\varepsilon\To{\fout} A$, where $A_\varepsilon(\ell)\coloneqq A(\ell+\varepsilon)$ is the sheaf of $\varepsilon$-extended behaviors, $\fin(a)=\rest{a}{0}{\ell}$, and $\fout(a)=\rest{a}{\varepsilon}{\ell+\varepsilon}$.
\end{example}

\begin{definition}[Composition of machines]
\label{def:composition}
Given two machines $M_1=(D_1,\fin_1,\fout_1)$ and $M_2=(D_2,\fin_2,\fout_2)$ of types $(A,B)$ and $(B,C)$ respectively, their \emph{composite} is the machine $M=(D_1\times_B D_2, \fin, \fout)$ of type $(A,C)$, namely the span given by pullback:
\begin{center}
\input{chapters/tikz/30_composition_event_based_span}
\end{center}
\end{definition}

Once we demand all our machines to be inertial, they do not form a category because there is no identity machine. However they do form an algebra on an operad of wiring diagrams; see \cite{Schultz2019} for details.

%

%% file: chapters/tikz/20_machine.tex
\begin{tikzcd}
  & C \arrow[ld, swap,"f^\text{in}"] \arrow[rd, "f^\text{out}"] &   \\
A &                         & B
\end{tikzcd}

%% file: chapters/tikz/20_cds.tex
\begin{tikzcd}
&D\arrow[swap]{dl}{\fin}\arrow{dr}{\fout}&\\
C^1(A)&&C^1(B)
\end{tikzcd}

%% file: chapters/tikz/30_composition_event_based_span.tex
%
\begin{tikzcd}
&&D_1 \times_B D_2 \arrow[ld, swap] \arrow[rd] \rotatedpullback  && \\
  & D_1 \arrow[ld, swap,"\fin_1"] \arrow[rd, "\fout_1"] & & D_2 \arrow[ld, swap,"\fin_2"] \arrow[rd, "\fout_2"]   \\
A& & B && C
\end{tikzcd}

%% file: chapters/event.tex
\section{Event-based Systems}
\label{sec:eventbased}

\begin{definition}[Event stream]
Let $A$ be a set and $\ell\geq 0$. We define a length-$\ell$ \emph{event stream} of type $A$ to be an element of the set
\begin{equation*}
    \event{A}(\ell)\coloneqq \{ (S,a) \mid S\subseteq \tilde{\ell}, S\text{ finite }, a\colon S\to A\}.
\end{equation*}
For an event stream $(S,a)$ we refer to elements of $S=\{s_1,\ldots,s_n\}\ss\tilde{\ell}$ as \emph{time-stamps}, we refer to $a$ as the \emph{event map}, and for each time-stamp $s_i$ we refer to $a(s_i)\in A$ as its \emph{value}.

If the set of time-stamps is empty, there is a unique event map, and we refer to $(\varnothing,!)$ as an \emph{empty event stream}.
\end{definition}

\begin{example}
Consider a Swiss traffic light and its set of color transitions 
\begin{equation*}
    A=\{\mathsf{redToOrange}, \mathsf{orangeToGreen},\mathsf{greenToOrange},\mathsf{orangeToRed}\}.
\end{equation*}
We will give an example of an element $(S,a)\in \event{A}(60)$, i.e.\ a possible event stream of length $60$. Its set of time-stamps is $S=\{20,25,45,50\}$, and the event map is given by
\begin{equation*}
    \begin{split}
        a\colon S&\to A\\
        s&\mapsto \begin{cases}
        \mathsf{redToOrange}, &\text{if }s=20,\\
        \mathsf{orangeToGreen}, &\text{if }s=25,\\
        \mathsf{greenToOrange}, &\text{if }s=45,\\
        \mathsf{orangeToRed}, &\text{if }s=50.
        \end{cases}
    \end{split}
\end{equation*}
\end{example}

\begin{definition}[Restriction map on event streams]\label{def.rest_event_stream}
For any event $e= (S\ss\tilde{\ell}, a\colon S\to A)\in \event{A}(\ell)$ and any $0\leq t\leq t'\leq \ell$, let $S_{t,t'}\coloneqq \{1\leq i \leq n \mid t\leq s_i \leq t'\}\ss S$, and let $a_{t,t'}\colon S_{t,t'}\to S\to A$ be the composite. Then we define the \emph{restriction} of $e$ along $[t,t']\subseteq \tilde{\ell}$ to be $\rest{e}{t}{t'}\coloneqq(S_{t,t'},a_{t,t'})$. 
\end{definition}

\begin{proposition}
\label{prop:evfunctorial}
$\event{}$ is functorial: Given a function $f\colon A\to B$ there is an induced morphism $\event{f}:\event{A}\to\event{B}$ in $\sheaf{\Int}$, and this assignment preserves identities and composition.
\end{proposition}

\begin{proposition}
\label{prop:evsheaf}
For any set $A$, the presheaf $\event{A}$ is in fact a sheaf.
\end{proposition}

\begin{remark}
\label{remark:monoidalstructure}
There is a monoidal structure $(\varnothing,\odot)$ on $\set$:  Its unit is $\varnothing$ and the monoidal product of $A$ and $B$ is $A\odot B\coloneqq A+B+A\times B$.
\end{remark}

\begin{proposition}
\label{prop:evstrong}
$\event{}\colon(\set,\odot,\varnothing)\to(\sheaf{\Int},\times,1)$ is a strong monoidal functor: $1\cong \event{\varnothing}$ and for any sets $A,B$, we have
\begin{equation*}
\event{A}\times\event{B}\cong\event{A\odot B}.
\end{equation*}
\proofref{app:evstrong}
\end{proposition}

\begin{definition}[Event-based system]
\label{def:eventbasedsystem}
Let $A,B$ be sets. An \emph{event-based system} $P=(C,\fin,\fout)$ of type  $(A,B)$ is a machine
\begin{equation}\label{eqn.span_ebs}
\input{chapters/tikz/30_event_based_span}
\end{equation}
between two event streams. For any input event stream $e\in \event{A}(\ell)$, the preimage $\left( \fin\right)^{-1}(e)$ is the set of all internal behaviors consistent with $e$.
\end{definition}

\begin{remark}
     An event-based system $P$ of type  $(A,B)$ is graphically represented as
\begin{equation*}
\input{chapters/tikz/30_event_based_system}
\end{equation*}
\end{remark}

%

\begin{example}[Discrete dynamical systems as event-based systems]
\label{ex:dds}
Let $A,B$ be sets. Then an \emph{$(A,B)$-discrete dynamical system (DDS)} consists of:
\begin{itemize}
    \item A set $S$, elements of which are called \emph{states}.
    \item A function $f\upd\colon A\times S\to S$, called the \emph{update function}.
    \item A function $f\rdt\colon S\to B$, called the \emph{readout function}.
\end{itemize}
We can transform any $(A,B)$ dynamical system into an $(A,B)$-event-based system as follows. Define $D$ to be the sheaf with the following sections:
\begin{equation*}
    D(\ell)\coloneqq\{T\ss\tilde{\ell}, (a,s)\colon T\to A\times S\mid T \tn{ finite and } s_{i+1}=f\upd(a_i,s_i)\tn{ for all }1\leq i\leq n-1\}.
\end{equation*}
The restriction map is the same as for the underlying event stream (\cref{def.rest_event_stream}).
We define the span \eqref{eqn.span_ebs} as follows. Given $(T,a,s)\in D(\ell)$, we have
 $\fin(T,a,s)=(T,a)$ and $\fout(T,a,s)=(T,(s\then f\rdt))$.
\end{example}

Given a function $f: S\to A$ and a subset $A'\ss A$, we can take the preimage $f\inv(A')\ss S$ and get a function $f\big|_{A'}\colon f\inv(A')\to A'$. This can be used to define a filter for event-based systems.

\begin{definition}[Filter for event-based systems]
\label{def:filter}
Let $A$ and $A'\subseteq A$ be sets. An $(A,A')$-\emph{filter} is an $(A,A')$-event-based system with $D=A$, $\fin=\identity{}$, and $\fout(\ell):\event{A}\to\event{A'}$ defined on $(S,a)$, where $S\ss\tilde{\ell}$ and $a\colon S\to A$, as follows:
\begin{equation*}
\fout(\ell)(S,a)\coloneqq( a\inv(A'),a\big|_{A'}).
\end{equation*}
In other words it consists of the subset (the preimage of $a\inv(A')\ss S$) of those time-stamps whose associated values are in $A'$, together with the original event map on that subset.
\end{definition}

\begin{definition}[Continuous stream]\label{def.continuous}
Let $A$ be a topological space. We define a \emph{continuous stream} of type $A$ to be
\begin{equation*}
    \cont{A}(\ell)\coloneqq \{a\mid a \colon \tilde{\ell}\to A \text{ continuous}\}.
\end{equation*}
\end{definition}
\begin{definition}[Lipschitz continuous function]
Given two metric spaces $(A,d_A)$ and $(B,d_B)$, a function $f\colon A\to B$ is called \emph{Lipschitz continuous} if there exists a $K\in \mathbb{R}$, $K\geq 0$, such that for all $a_1,a_2\in A$:
\begin{equation*}
    d_B(f(a_1),f(a_2))\leq Kd_A(a_1,a_2).
\end{equation*}
\end{definition}
\begin{remark}[Lipschitz continuous stream]
In \cref{def.continuous}, if $A$ is a metric space, then we can consider the subsheaf consisting of only those streams $a\colon\tilde\ell\to A$ that are Lipschitz continuous, i.e.
\begin{equation*}
\lcont{A}(\ell)=\{a\colon\tilde\ell\to A\mid a\tn{ Lipschitz continuous}\}\ss\cont{A}(\ell).
\end{equation*}
\end{remark}

\begin{example}
For any $S$, there is a \emph{codiscrete} topological space $\hat{S}$ with points $S$ and only two open sets: $\varnothing$ and $S$. For any topological space $X$, the functions from the underlying set of $X$ to $S$ are the same as the continuous maps $X\to\hat{S}$.%
\footnote{Technically, one can say that the underlying set functor is left adjoint to the codiscrete functor.}
Thus, we have
\begin{align*}
    \cont{\hat{S}}(\ell)&=
    \{a\mid a\colon\tilde{\ell}\to\hat{S}, a \text{ continuous}\}\\&=
    \{a\mid a\colon\tilde{\ell}\to S\}.
\end{align*}
We sometimes denote $\cont{\hat{S}}$ simply by $\cont{S}$.
\end{example}

\begin{definition}[Sampler]\label{def.sampler}
Let $A$ be a topological space and choose $d\in \Rgeq$, called the \emph{sampling time}. A \emph{period-$d$ $A$-sampler} is a span
\begin{center}
\input{chapters/tikz/30_sampler}
\end{center}
where $\fcnt,\fevt$ are morphisms:
\begin{equation*}
    \begin{split}
        \fcnt \colon \clock{d} \times\cont{A}&\to \cont{A}\\
        (\phi,a)&\mapsto a,\\
        \fevt \colon \clock{d} \times\cont{A}&\rightarrow \event{A}\\
        (\phi,a)&\mapsto\phi\then a.
    \end{split}
\end{equation*}
The second formula is the composite $\phi\subseteq \tilde{l}\To{a}A$, which takes the value of $a$ at $d$-spaced intervals. 
\end{definition}

\begin{definition}[$L$-level-crossing sampler]
\label{def:levelcrossing}
Let $(A,\mathrm{dist})$ be a metric space and consider a Lipschitz input stream $\lcont{A}(\ell)$. Consider the \emph{level} $L\in \rr$. A \emph{$L$-level-crossing sampler} of type $A$ is a machine
\begin{center}
\input{chapters/tikz/30_level_crossing_span}
\end{center}
with $P(\ell)\coloneqq \{(c,a_0) \mid c\in \lcont{A}(\ell), a_0\in A\}$. Given $p=(c,a_0)\in P(\ell)$, we have $\fcnt(c,a_0)=c$.
Furthermore, either $\mathrm{dist}(c(t),a_0)<L$ for all $t \in \tilde
{\ell}$ or there exists $t\in \tilde{\ell}$ with $\mathrm{dist}(c(t_1),a_0) \geq L$. In the first case, take $\fevt(c)=(\varnothing,!)$ to be the empty event stream. In the second case, define
\begin{equation*}
    t_1=\inf\{t\in \tilde{\ell}\mid\mathrm{dist}(c(t_1),a_0) \geq L\}
\end{equation*}
and let $a_1=c(t_1)$. Recursively, define $t_{i+1}\in \tilde{\ell}$ to be the least time such that $\mathrm{dist}(c(t_{i+1}),a_i) \geq L$ (if there is one). There will be a finite number of these because $c$ is Lipschitz. We denote the last of such times by $t_n$. Then, we define
\begin{equation*}
    \begin{split}
        \fevt\colon P(\ell) &\to \event{A}\\
        (c,a_0) &\mapsto \{t_1,\ldots,t_n,c(t_1),\ldots,c(t_n)\}.
    \end{split}
\end{equation*}
\end{definition}

\begin{definition}[Reconstructor]
\label{def:reconstructor}
Let $A$ be a set. A \emph{reconstructor} of type $A$ is a span
\begin{center}
\input{chapters/tikz/30_reconstructor}
\end{center}
with $C(\ell)=\{(S,a_0,a)\mid S\ss\tilde{\ell}\tn{ finite}, a_0\in A, a\colon S\to A\}$, where $\fevt\colon C\to\event{A}$ is given by $
\fevt(S,a_0,a)\coloneqq (S,a)$,
and where $a'\coloneqq\fcnt(a_0,s_1,\ldots,s_n,a_1,\ldots,a_n)\colon\tilde{\ell}\to A$ is given by
\begin{equation*}
    a'(t)\coloneqq
    \begin{cases}
    a_0,& 0\leq t<s_1\\
    a(s_i),&s_i\leq t <s_{i+1}, \quad i\in \{2,\ldots,n-1\}\\
    a(s_n),&s_n\leq t\leq \ell.
    \end{cases}
\end{equation*}
\end{definition}

\begin{remark}
The reconstructed stream is not continuous, it is only piecewise continuous. Luckily, it is continuous with respect to the codiscrete topology. We denote $\hat{A}$ simply by $A$.
\end{remark}
\begin{remark}
The reconstructor is known in signal theory as the zero-order-hold (ZOH), and represents the practical signal reconstruction performed by a conventional digital-to-analog converter (DAC).
\end{remark}

\begin{definition}[Composition of event-based systems]
\label{def:compositionevent}
Given two event-based systems $P_1=(D_1,\fin_1,\fout_1)$ and $P_2=(D_2,\fin_2,\fout_2)$ of types $(A,B)$ and $(B,C)$, they compose as machines do (\cref{def:composition}). Their \emph{composite} is the $(A,C)$-event-based system $P=(D_1\times_B D_2, \fin, \fout)$.
\end{definition}

\begin{remark}
    We refer to the composition of event-based systems as putting them in \emph{series}, and represent it graphically as
\begin{equation*}
\input{chapters/tikz/30_composition_event_based}
\end{equation*}
\end{remark}

%

\begin{definition}[Tensor Product of event-based systems]
\label{def:product}
Given two event-based systems $P_1=(E_1,\fin_1,\fout_1)$ and $P_2=(E_2,\fin_2,\fout_2)$ of types $(A,B)$ and $(C,D)$, their tensor product is an event-based system $P=(E_1\times E_2, \fin, \fout)$ of type $(A\odot C, B\odot D)$, i.e. a span
\begin{center}
\input{chapters/tikz/30_product_event_based_span}
\end{center}
This is an event-based system because $\event{A}\times\event{C}\cong\event{A\odot C}$ and $\event{B}\times\event{D}\cong\event{B\odot D}$.
\end{definition}
\begin{remark}
    We refer to the tensor product of event-based systems as putting them in \emph{parallel}, and represent it graphically as
    \begin{equation*}
    \input{chapters/tikz/30_product_event_based}
    \end{equation*}
\end{remark}

\begin{definition}[Trace of event-based system]
\label{def:trace}
Given an event-based system $P=(D,\fin,\fout)$ of type $(A\times C, B\times C)$, its trace is an event-based system $P'=(E_{\mathrm{tr}},\fin_{\mathrm{tr}}, \fout_{\mathrm{tr}})$ of type $(A,B)$ given by 
\[
E_{\mathrm{tr}}(\ell)=\{d\in D(\ell)\mid\pi_2(\fin(d))=\pi_2(\fout(d))\}
\]
with $\fin_{\mathrm{tr}}(d)=\pi_1(\fin(d))\in\event{A}$ and $\fout_{\mathrm{tr}}(d)=\pi_1(\fout(d))\in\event{B}$.
\end{definition}

\begin{remark}
We depict an event-based system with trace as having \emph{feedback} or a \emph{loop}, and represent it graphically as
\begin{equation*}
\input{chapters/tikz/30_trace_diagram}
\end{equation*}
\end{remark}

%% file: chapters/tikz/30_event_based_span.tex
\begin{tikzcd}
  & C \arrow[ld, swap,"\fin"] \arrow[rd, "\fout"] &   \\
\event{A} &                         & \event{B}
\end{tikzcd}

%% file: chapters/tikz/30_event_based_system.tex
\begin{tikzpicture}[oriented WD, bb min width =1.5cm, bby=2ex, bbx=.7cm,bb port length=3pt]
    \node[bb port sep=0.8, bb={1}{1}, bb name={$P$}] (ev) {};
    \node [black, left = 0.2 of ev] {$\event{A}$};
    \node [black, right = 0.2 of ev] {$\event{B}$};
\end{tikzpicture}

%% file: chapters/tikz/30_sampler.tex
\begin{tikzcd}
  & \clock{d}\times\cont{A}\arrow[ld, swap,"\fcnt"] \arrow[rd, "\fevt"] &   \\
\cont{A} &                         & \event{A}
\end{tikzcd}

%% file: chapters/tikz/30_level_crossing_span.tex
\begin{tikzcd}
    &P\arrow[swap]{dl}{\fcnt}\arrow{dr}{\fevt}&\\
    \lcont{A}&&\event{A}
\end{tikzcd}

%% file: chapters/tikz/30_reconstructor.tex
\begin{tikzcd}
  & C \arrow[ld, swap,"\fevt"] \arrow[rd, "\fcnt"] &   \\
\event{A} &                         & \cont{A}
\end{tikzcd}

%% file: chapters/tikz/30_composition_event_based.tex
\begin{tikzpicture}[oriented WD, bb min width =1.5cm, bby=2ex, bbx=.7cm,bb port length=3pt]
    \node[bb port sep=0.8, bb={1}{1}, bb name=${P_1}$] (p1) {};
    \node[bb port sep=0.8, bb={1}{1},right=2 of p1.east, bb name=${P_2}$] (p2) {};
    \node[bb port sep=0.8, bb={1}{1},right=4 of p2.east, bb name=${P}$] (p3) {};
    \node [black, left = 0.2 of p1] {$\event{A}$};
    \draw[ar] (p1) to node[pos=0.5,above] {$\event{B}$} (p2);
    \node [black, right = 0.2 of p2] {$\event{C}$};
    \node at ($(p2.east)!.5!(p3.west)$) {$\equiv$};
    \node [black, left = 0.2 of p3] {$\event{A}$};
    \node [black, right = 0.2 of p3] {$\event{C}$};
\end{tikzpicture}

%% file: chapters/tikz/30_product_event_based_span.tex
\begin{tikzcd}
  & E_1\times E_2 \arrow[ld, swap,"\fin_1\times\fin_2"] \arrow[rd, "\fout_1\times\fout_2"] &   \\
\event{A}\times\event{C} &                         & \event{B}\times\event{D}
\end{tikzcd}

%% file: chapters/tikz/30_product_event_based.tex
    \begin{tikzpicture}[oriented WD, bb min width =1.5cm, bby=2ex, bbx=.7cm,bb port length=3pt]
    \node[bb port sep=0.8, bb={1}{1}, bb name=${P_1}$] (dp_1) {};
    \node[bb port sep=0.8, bb={1}{1},below=1.5 of dp_1.south, bb name=${P_2}$] (dp_2) {};
    \node[bb port sep=0.8, bb={1}{1},below right=0.75 and 5.5 of dp_1.east, bb name=${P}$] (dp_3) {};
    \node [black, left = 0.2 of dp_1] {$\event{A}$};
    \node [black, right = 0.2 of dp_1] {$\event{B}$};
    \node [black, left = 0.2 of dp_2] {$\event{C}$};
    \node [black, right = 0.2 of dp_2] {$\event{D}$};
    \coordinate (helper) at (dp_3-|dp_1.east);
    \node at ($(helper)!.5!(dp_3.west)$) {$\equiv$};
    \node [black, left = 0.2 of dp_3] {$\event{A\odot C}$};
    \node [black, right = 0.2 of dp_3] {$\event{B\odot D}$};
    \end{tikzpicture}

%% file: chapters/tikz/30_trace_diagram.tex
\begin{tikzpicture}[oriented WD, bb min width =1.5cm, bby=2ex, bbx=.7cm,bb port length=3pt]
    \node[bb port sep=0.8, bb={2}{2}, bb name={$P$}] (Loop) {};
    \draw[ar] let \p1=(Loop.north east), \p2=(Loop.north west), \n1={\y1+\bby}, \n2=\bbportlen in (Loop_out1) to[in=0] (\x1+\n2,\n1) -- (\x2-\n2,\n1) to[out=180] (Loop_input1);
    \node [black, above right = 0.1 and 0.2 of Loop] {$\event{C}$};
    \node [black, below left = 0.0 and 0.2 of Loop.west] {$\event{A}$};
    \node [black, below right = 0.0 and 0.2 of Loop.east] (evB) {$\event{B}$};
    \node[bb port sep=0.8, bb={1}{1}, bb name={$P'$}, right = 4.5 of Loop] (noloop) {};
    \node [black, right = 0.2 of noloop] {$\event{B}$};
    \node [black, left = 0.2 of noloop] (evA) {$\event{A}$};
    \node at ($(evB)!.5!(evA)$) {$\equiv$};
\end{tikzpicture}

%% file: chapters/examples.tex
\section{Example: Neuromorphic Optomotor Heading Regulation}
In the following, we want to show that complex \glspl{abk:cps} can be modeled using the framework presented in Section~\ref{sec:eventbased}. 
To do so, we consider the neuromorphic optomotor heading regulation problem studied in~\cite{Censi2015}. Specifically, we consider the case of a body moving in the plane and changing its orientation, expressed as an element of SO(2), based on the scene observed by an event camera mounted on it. The result is reported in Figure~\ref{fig:robot}, which shows that this complex system can be understood as the composition (\cref{def:composition}) of machines.
\label{sec:examples}

\subsection{Background on Event Cameras}
\label{sec:backgroundevent}
Event cameras are \emph{asynchronous} sensors which have introduced a completely new acquisition technique for visual information~\cite{Licht2008}.  This new type of sensors samples light depending on the scene dynamics and therefore differs from standard cameras. In particular, event cameras show notable advantages such as high temporal resolution, low latency (in the order of microseconds), low power consumption, and high dynamic range, all of which naturally encourage their employment in robotic applications. An exhaustive review of the existing applications of event cameras has been reported in~\cite{Gallego2019}.

\begin{figure}[tb]
\begin{center}
\subfloat[\label{fig:dvsa}]{\includegraphics[width=0.31\columnwidth]{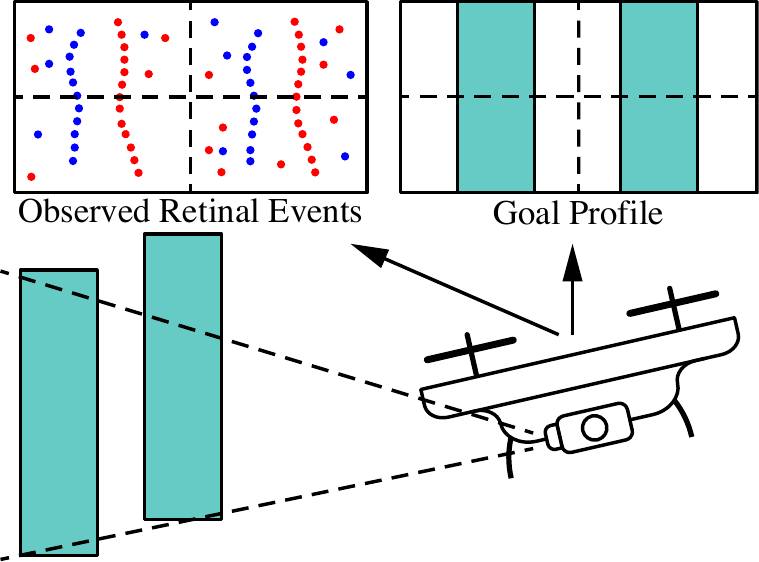}}
\quad
\subfloat[\label{fig:dvsb}]{\includegraphics[width=0.31\columnwidth]{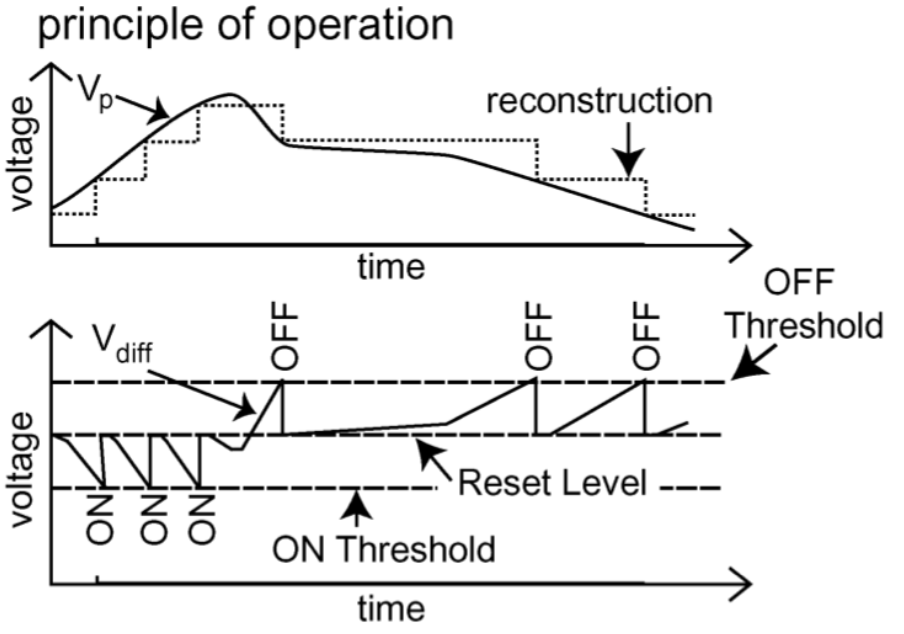}}
\quad
\subfloat[\label{fig:dvsc}]{\includegraphics[width=0.31\columnwidth]{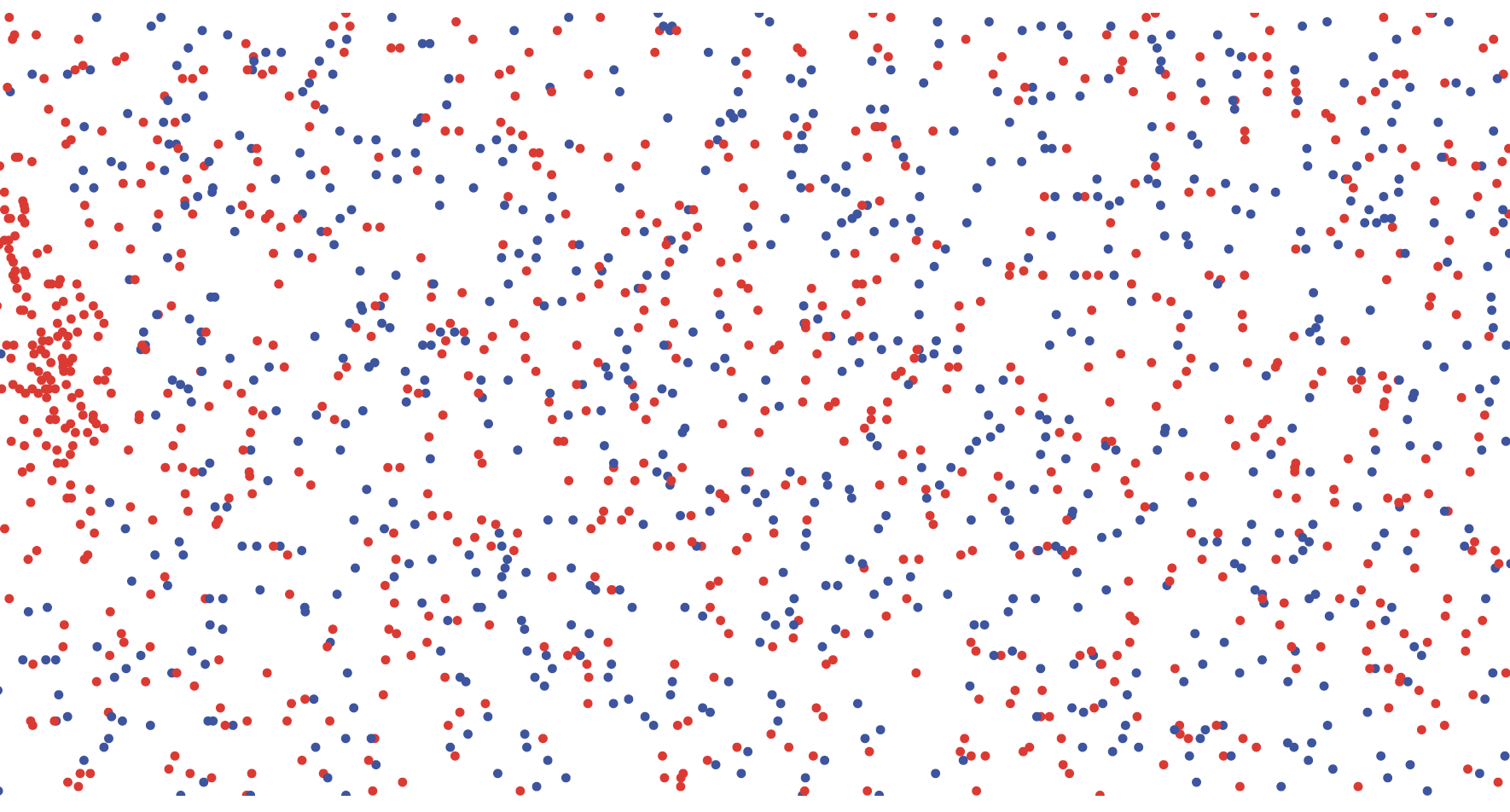}}
\end{center}
\caption{(a) Illustration of  the neuromorphic heading regulation problem. (b) Working principle of the Dynamic Vision Sensor~\cite{Licht2008}. (c) Events over a given time-window~\cite{Censi2015} \label{fig:dvs}.}
\end{figure}

\subsubsection*{Event Generation Model}
In the following, we review the event generation model presented in~\cite{Censi2015}. An event camera~\cite{Licht2008} is composed of a set $\mathcal{S}$, elements of which are called pixels, reacting (independently) to changes in light \emph{brightness}. There is a function $\dir \colon \mathcal{S}\to \mathbb{S}^1$, representing the \emph{direction} of each pixel in the event camera's field of view. The environment reflectance is a function $m\colon \mathbb{S}^1 \to \Rgeq$, such that $m(\dir(s))$ represents the intensity of light from direction $\dir(s)\in \mathbb{S}^1$ at any given moment of time. The light field is a map
\begin{equation}
\label{eq:lightInt}
\begin{split}
    I\colon \Rgeq \times \mathcal{S} &\to \Rgeq\\
    \Pair{t}{s} &\mapsto I_{t}^{s},
\end{split}
\end{equation}
where $I_{t}^{s}$ represents the intensity of light reaching the event camera at time $t$ in direction $\dir(s)\in \mathbb{S}^1$. Brightness is expressed as $L\coloneqq \log(I_{t}^{s})$. An event $e=\triple{s}{t}{p}$ is generated at pixel location $s$ at time $t$ if the change 
\begin{equation*}
    \Delta L(s,t)\coloneqq L(s,t) - L(s, t')
\end{equation*}
in brightness at that pixel since the last event $t'$ was fired reaches a threshold $\pm C$ (Figure~\ref{fig:dvsb}), i.e. $|\Delta L(s,t)|= pC$, where $p\in \{-1,1\}$ represents the event's polarity~\cite{Gallego2019} (\textcolor{blue}{blue} and \textcolor{red}{red} in Figure~\ref{fig:dvs}).

\begin{remark}
Note that the contrast sensitivity $C$ can be tuned depending on the applications through the pixel bias currents, as explained in~\cite{Nozaki2017}.
\end{remark}
\subsection{Robotic System Description}
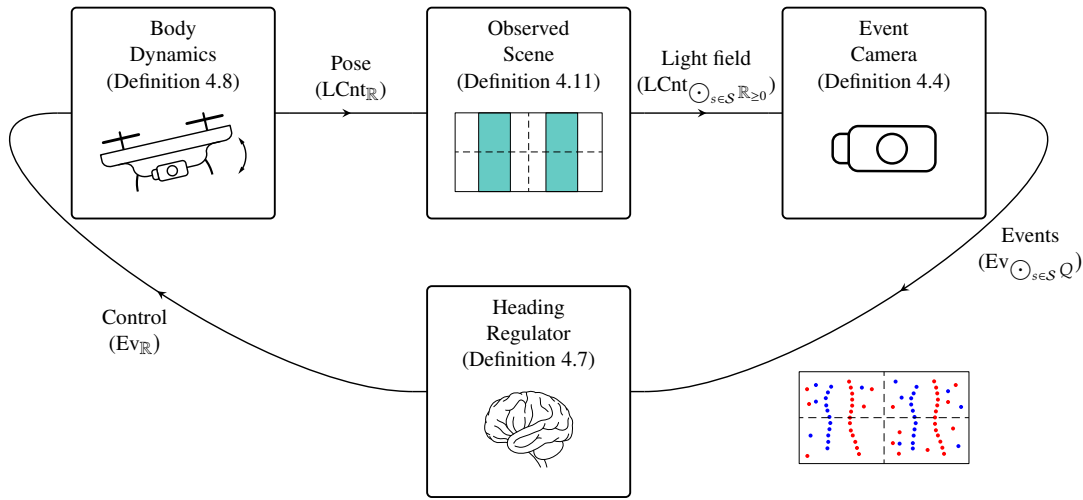
\begin{figure}[tbh]
\begin{center}
\input{chapters/tikz/40_closed_loop}
\caption{Graphical representation of the neuromorphic heading regulation problem as a composition of machines, together with input-output stream types.}
\label{fig:robot}
\end{center}
\end{figure}

In this section, we describe the robotic system presented in~\cite{Censi2015}. The system is composed of a body, which is able to move on the plane and to change its orientation (heading), expressed as an element of SO(2). An event camera is mounted on the body, and allows for it to perceive the environment. Furthermore, heading regulation happens in image space via feedback from the event camera through a decision process (regulator). A graphical representation of the robotic system is shown in Figure \ref{fig:robot}.

In the following, each one of the aforementioned components will be described as a machine, showing the unifying properties of our framework. The components' interactions will be represented through composition and product of machines. The diagram reported in Figure~\ref{fig:robot} contextualizes the following definitions.

\subsubsection{Event Camera}
To represent an event camera in our framework, we first consider a single pixel as a machine.  In \cref{def:eventcamera} we will define an event camera to be the product (Definition~\ref{def:product}) of $n$ independent event camera pixels.  

\begin{definition}[Event camera pixel]
\label{def:ebpixel}
First, define a $(\lcont{\Rgeq},\lcont{\rr})$ machine $P_1$
\begin{center}
\begin{tikzcd}
&A\arrow[dl,equal,swap]\arrow{dr}{\fout}&\\
\lcont{\Rgeq}&&\lcont{\rr}
\end{tikzcd}
\end{center}
with $A(\ell)\coloneqq \{a \colon \tilde{\ell}\to \rr\mid a\text{ Lipschitz continuous}\}.$ For $a\in A(\ell)$, define $\fin(a)=a$ (i.e. the identity, from now on as in the given span) and $\fout(a)=\log(a)$. Then,
for any $C\in \Rgeq$, called the \emph{contrast sensitivity}, let $P_2$ be a $C$-level-crossing sampler (\cref{def:levelcrossing}) of type $\rr$, with input $\lcont{\rr}$ and output $\event{\rr}$. Finally, let $P_3$ be the $(\rr,Q)$-event-based system corresponding to the DDS (\cref{ex:dds}) of input-output type $(\rr,Q)$, $Q=\{-1,1\}$, with state set $S=\Rgeq \times Q$, consisting of pairs $(r,q)$, readout $f\rdt\colon S\to Q$ defined as $f\rdt(r,q)=q$, and update function defined as
\begin{equation*}
    \begin{split}
        f\upd\colon \rr\times S&\to S\\
        (r',r,q)&\mapsto
        \begin{cases}
        (r',1),& r'-r\geq C\\
        (r',-1),&r'-r\leq C.
        \end{cases}
    \end{split}
\end{equation*}
 
An \emph{event-camera pixel} for an event camera with contrast sensitivity $C$ is the composite (\cref{def:composition}) machine $P_C=P_1\then P_2\then P_3$ with input of type $\lcont{\Rgeq}$ and output of type $\event{Q}$.
\end{definition}

\begin{remark}[Interpretation]
Recalling the working principle of an event camera pixel, the continuous input of $P^C$ represents the light intensities measured by a pixel at a specific time (Equation~\ref{eq:lightInt}). The machine $P_1$ computes the $\log$ of the intensities, and the $C$-level-crossing sampler $P_2$ determines when the changes in intensities are sufficient to generate events. The DDS $P_3$ determines the polarity of the events, by storing it in the state set together with the brightness of the previously fired event.
\end{remark}
We are now ready to define an event camera as a machine, arising from the product of the event camera pixels (machines) composing it.
\begin{definition}[Event camera]
\label{def:eventcamera}
Let $\cat{S}$ be a set, elements of which we think of as pixels. For any contrast sensitivity $C\in\Rgeq$, define the \emph{event camera with $\cat{S}$ pixels and contrast sensitivity $C$} to be the product (Definition~\ref{def:product}) $\prod_{s\in\cat{S}}P_C=(P_C)^\mathcal{S}$, where $P_C$ is as in \cref{def:ebpixel}. The input stream of $P_C$ is of type $\lcont{\bigodot_{s\in \cat{S}}\Rgeq}$ and the output stream is of type $\event{\bigodot_{s\in \cat{S}}Q}$.
\end{definition}

\begin{remark}
Recall the monoidal structure presented in \cref{remark:monoidalstructure}. The strong monoidality of $\event{}$ (\cref{prop:evstrong}) means that each event of the event camera consists of an event at one or more of its pixels.
\end{remark}

\subsubsection{Heading Regulator}
Given the body with heading $\theta_t$, one wants to steer it toward some ``goal'' heading $\theta_g$, i.e. one wants to reach $\theta_{t+\Delta_t} \in [\theta_g-\delta,\theta_g+\delta]$, for some probability $1-\varepsilon(\delta)$. This is achieved by a \emph{heading regulator}, which only considers the events measured by the event camera, and takes decisions in real time. To do so, one needs a function $f:\mathcal{S}\rightarrow \rr$, called the \emph{estimator}, which for each event observed on pixel $s_i \in \mathcal{S}$ at time $t_j$ gives an estimate of the current heading of the body $\theta_{t_j}=f(s_i)$. As shown in~\cite{Censi2015}, it is sufficient to find an $f$ such that
\begin{equation}
\label{eq:functionForStats}
    \int_{\mathcal{S}}f(s)p_e(s,t)\text{d}s=\theta_t,
\end{equation}
in a neighborhood of $\theta_g$, where $p_e(s,t)$ represents the probability of event $e$ being generated at pixel $s$ at time $t$.
\begin{remark}
As shown in~\cite{Censi2015}, a possible choice for $f$ is
\begin{equation*}
\begin{split}
    f:\mathcal{S}&\rightarrow \rr\\
    s&\mapsto f(s)\coloneqq \int_{s-\delta}^{s+\delta}p_{\theta}(s-v)\vert \nabla m^v\vert (s-v)\text{d}v,
\end{split}
\end{equation*}
where each pixel $v$ contributes with $(s-v)$, weighted by the probability of generating an event $(\nabla m^v)$, given the probability $p_\theta(s-v)$ of $\theta$ being $s-v$.
\end{remark}
The events used to regulate the heading are described through statistics $S_t\in \rr$, which are computed asynchronously, when an event is observed. Given a function $f$, one can define the heading regulator as an event-based system.

Note that $\bigodot_{s\in\cat{S}}Q\cong\{(S,q)\mid S\ss\cat{S} \text{ non-empty}, q\colon S\to Q\}$. In fact, the heading regulator will not use the polarities $q$, but only the firing sets $S$. We will follow the style of \cref{ex:dds}, but what follows does not arise from a DDS because we explicitly use the time-stamps.

\begin{definition}[Heading regulator]
\label{def:headingregulatorEB}
Given an event-camera (\cref{def:eventcamera}), we define an $(\bigodot_{s\in \cat{S}}Q,\rr)$ event-based system $H=(D,\fin,\fout)$ (the \emph{heading regulator}) as follows. Let $X=\Rgeq\times\rr$ and define
\begin{equation*}
    D(\ell)\coloneqq \big\{t_1,\ldots,t_n, (S_1,q_1),\ldots,(S_n,q_n),x_1,\ldots,x_n
    \mid x_i=f\upd((S_i,q_i),t_i,x_{i-1})\big\},
\end{equation*}
where $\{t_1,\ldots,t_n\} \subseteq \tilde{\ell}$, $x_i\in X$, $(S_i,q_i)\in \bigodot_{s\in S}Q$, and
\begin{equation*}
\begin{split}
    f\upd \colon  \textstyle\bigodot_{s\in \cat{S}}Q\times \Rgeq \times X&\to X\\
    ((S,q),t,x) &\mapsto \left(t,\sum_{s_i\in S} e^{-a(t-\pi_1(x))}\pi_2(x)-\frac{\kappa}{a}f(\dir(s_i))\right),
\end{split}
\end{equation*}
with $f$ satisfying Equation~\ref{eq:functionForStats}, and $a>0,\kappa>0$ tunable parameters. Then define $f\rdt(x)=\pi_2(x)$ and
\begin{equation*}
    \begin{split}
        \fin \colon D(\ell)&\to \event{\textstyle\bigodot_{s\in \cat{S}}Q}\\
        d&\mapsto \fin(d)\coloneqq \{t_1,\ldots,t_n,(S_1,q_1),\ldots,(S_n,q_n)\}, \quad (S_i,q_i)\in \textstyle\bigodot_{s\in \cat{S}}Q,
    \end{split}
\end{equation*}
and
\begin{equation*}
    \begin{split}
        \fout \colon D(\ell)&\to \event{\rr}\\
        d&\mapsto \fout(d)\coloneqq \{t_1,\ldots,t_n,f\rdt(x_1),\ldots,f\rdt(x_n)\}, \quad x_i\in X.
    \end{split}
\end{equation*}
\end{definition}

\subsubsection{Body Dynamics}
For small variations, the body orientation $\theta_t\in\rr$ and its dynamics are expressed through the law
\begin{equation*}
    \text{d}\theta_t=\mathsf{sat}_b(u)\text{d}t,
\end{equation*}
where $u\in \rr$ represents the input received from the controller and
\begin{equation*}
\begin{split}
    \mathsf{sat}_b\colon \rr&\to \rr_{[-b,b]}\\
    r&\mapsto \mathsf{sat}_b(r)\coloneqq \begin{cases}
    r,&\text{if }r\in [-b,b],\\
    -b,&\text{if }r<-b,\\
    b,&\text{if }r>b
    \end{cases}
\end{split}
\end{equation*}
represents the saturation of the actuators, which receiving a control $r\in \rr$, are only able to commute it to an actuation $\mathsf{sat}_b(r)\in [-b,b]$. The body dynamics can be written as a machine.
\begin{definition}[Body dynamics]
\label{def:body}
Consider a reconstructor (\cref{def:reconstructor}) $P_1$ of type $\rr$ with input $\event{\rr}$ and output $\cont{\rr}$. Furthermore, consider a continuous dynamical system $P_2$ (\cref{ex:cds}) of input-output type $(C^1(\rr),C^1(\rr))$, with state space $X=\rr$ representing the pose of the robot. Then, define the dynamics as $\dot{s}=\mathsf{sat}_{b}(u)$ and the readout as the identity, i.e. $f\rdt(s)=s$. The \emph{body dynamics} $P$ are given by the composition (\cref{def:composition}) of machines $P_1\then P_2$. It has input stream of type $\event{\rr}$ and output stream of type $\lcont{\rr}$.
\end{definition}
\begin{remark}[Interpretation]
The heading regulator produces an event stream of type $\event{\rr}$. However, the body dynamics are expressed in continuous time and therefore one needs a reconstructor $P_1$. Then, $P_2$ just represents the continuous dynamics.
\end{remark}

\subsubsection{Observed Scene}
Considering the variations in the dynamics, resulting in the current pose $\theta_t$, one can write the variations of the light intensities for each pixel $s\in \mathcal{S}$ with direction $\dir(s)\in\mathbb{S}^1$ as $I_{t}^{s}=m(\theta_t+\dir(s))$, where $m$ represents the environment reflectance introduced in Section~\ref{sec:backgroundevent}.

\begin{definition}[Scene observed by an event-camera pixel]
\label{def:scenepixel}
Given a pixel $s\in \mathcal{S}$ of an event-camera, let $\dir(s)$ be its fixed direction. The \emph{scene observed by an event-camera pixel} is an $(\lcont{\rr},\lcont{\Rgeq})$ machine
\begin{center}
    \begin{tikzcd}
        &\lcont{\rr}\ar[dl, equal]
        \ar[dr, "\fout"]&\\
        \lcont{\rr}&&\lcont{\Rgeq}
    \end{tikzcd}
\end{center}
For $\theta\in \lcont{\rr}(\ell)$, define
\begin{equation*}
    \begin{split}
        \fout\colon \lcont{\rr}(\ell)&\to \lcont{\Rgeq}\\
        \theta&\mapsto (\theta+\dir(s))\then m.
    \end{split}
\end{equation*}
\end{definition}

\begin{definition}[Scene observed by an event-camera]
\label{def:scenecamera}
Consider an event camera as in \cref{def:eventcamera}. The scene observed by each pixel of the camera is a machine $P$ of type $(\lcont{\rr},\lcont{\Rgeq})$ (Definition~\ref{def:scenepixel}).  The \emph{scene observed by an event camera} is the machine given by the product (Definition~\ref{def:product}) of machines $\prod_{s\in\cat{S}}P=P^\mathcal{S}$.
\end{definition}

Note that the output of the machine presented in Definition~\ref{def:scenecamera} is of the same type of the input of the machine presented in Definition~\ref{def:eventcamera}. This allows us to close the loop reported in Figure~\ref{fig:robot} using the trace (Definition~\ref{def:trace}). 

Consider an event camera $C$ (Definition~\ref{def:eventcamera}), a heading regulator $H$ (Definition~\ref{def:headingregulatorEB}), the body dynamics $B$ (Definition~\ref{def:body}), and the scene observed by the event camera $O$ (Definition~\ref{def:scenecamera}). In order to take the trace (and have the result be deterministic and total) we compose with a $\rr$-type $\varepsilon$-delay machine (\cref{ex:delay}) $\mathrm{Del}^\rr_\varepsilon$.  Then the trace
\begin{equation*}
    \mathrm{Tr}^{\rr}(B\then O\then C\then H \then \mathrm{Del}^{\rr}_\varepsilon),
\end{equation*}
 of the composite machine (Definition~\ref{def:composition}), is the desired closed-loop behavior of the robotic system.

%% file: chapters/tikz/40_closed_loop.tex
\scalebox{0.9}{
\hspace{-1.75cm}
\begin{tikzpicture}[oriented WD, bb min width =2.5cm, bby=2ex, bbx=.7cm,bb port length=3pt,font=\footnotesize]
\node[bb port sep=4.5pt, bb={1}{1},bb min width=3cm, bb name={\begin{tabular}{c} Body \\ Dynamics \\ (\cref{def:body})\\[+7pt] \includegraphics[scale=0.6]{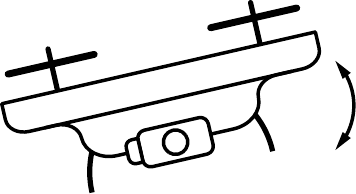}\end{tabular}}] (body) {};
\node[bb port sep=4.5pt,bb min width=3cm, bb={1}{1}, bb name={\begin{tabular}{c}Observed\\ Scene \\ (\cref{def:scenecamera})\\[+7pt] \includegraphics[scale=0.6]{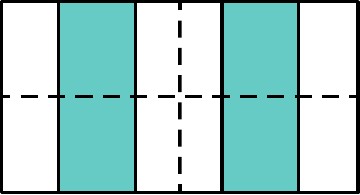}\end{tabular}},right=2.25cm of body] (scene) {};
\node[bb port sep=4.5pt,bb min width=3cm, bb={1}{1}, bb name={\begin{tabular}{c}Event \\ Camera \\(\cref{def:eventcamera})\\[+12pt] \includegraphics[scale=1.75]{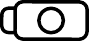}\end{tabular}}, right=2.25cm of scene] (camera) {};
\node[bb port sep=4.5pt, bb min width=3cm, bb={1}{1}, bb name={\begin{tabular}{c}Heading \\ Regulator \\(\cref{def:headingregulatorEB})\\[+7pt] \includegraphics[scale=0.6]{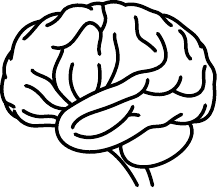}\end{tabular}}, below=1cm of scene] (controller) {};
\draw[ar] (body_out1) to node[pos=0.5,above] {\begin{tabular}{c}Pose \\ ($\lcont{\rr}$)\end{tabular}} (scene_input1);
\draw[ar] (scene_out1) to node[pos=0.5,above] {\begin{tabular}{c}Light field \\ ($\lcont{\bigodot_{s\in \cat{S}}\Rgeq}$)\end{tabular}}  (camera_input1);
\draw[ar] (camera_out1) to[out=0,in=0] node[pos=0.5,right] {\begin{tabular}{c}Events \\ ($\event{\bigodot_{s\in \cat{S}}Q})$\end{tabular}} (controller_out1);
\draw[ar] (controller_input1) to[out=180,in=180] node[pos=0.3,left=0.5cm] {\begin{tabular}{c}Control \\ ($\event{\rr})$\end{tabular}} (body_input1);
\node[inner sep=0pt] (eventview) at (15,-13) {\includegraphics[scale=0.7]{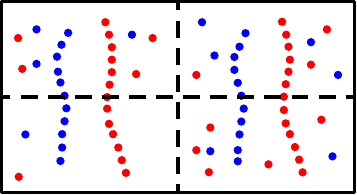}};
\end{tikzpicture}}

%% file: chapters/conclusion.tex
\section{Conclusion and future Work}
\label{sec:conclusion}
In this paper, we presented a framework characterized by high descriptive power and formality, and we showed how event-based systems can be modeled using it. However, the framework does not subsume the literature presented in \cref{sec:intro} yet. In particular, we look forward to exploring three extensions. First, we would like to explicitly introduce a notion of uncertainty, which would allow for a more accurate description of particular systems. Second, we would like to explore the implications of super-dense time, mentioned in~\cite{Lee2010}. Third, we aim at developing tools for the synthesis of event-based systems, using the internal language of the topos of behaviour types~\cite{Schultz2019}, and the mathematical theory of co-design 
~\cite{CensiCoDesign2015}.